\documentclass[runningheads]{llncs}
\usepackage{graphicx}
\usepackage{color}
\usepackage[ruled]{algorithm2e}
\usepackage[colorlinks,linkcolor=red]{hyperref}
\usepackage{url}
\usepackage{stfloats}
\usepackage{multirow}
\usepackage{makecell}
\usepackage{float}
\usepackage{bbding}
\usepackage[misc]{ifsym}

\begin{document}
\title{Spirit Distillation: A Model Compression Method with Multi-domain Knowledge Transfer\thanks{Supported by the National Natural Science Foundation of China under Grant 62072211, 51939003, U20A20285. Full version of our article is available at \url{arxiv.org/pdf/2103.13733.pdf}.}}
\titlerunning{Spirit Distillation}
%

\author{Zhiyuan Wu\inst{1} \and
Yu Jiang\inst{1,2} \and
Minghao Zhao\inst{1} \and
Chupeng Cui\inst{1} \and \\
Zongmin Yang\inst{1} \and 
Xinhui Xue\inst{1} \and
Hong Qi\inst{1,2}\textsuperscript{(\Letter)}
}
\authorrunning{Z. Wu et al.}
%
\institute{College of Computer Science and Technology, Jilin University, Changchun, China \and
Key Laboratory of Symbolic Computation and Knowledge Engineering of Ministry of Education, Jilin University, Changchun, China \\
\email{\{qihong,jiangyu2011\}@jlu.edu.cn}\\
\email{\{wuzy2118,zhaomh20,cuicp2118,yangzm2118,xuexh2118\}@mails.jlu.edu.cn}
}
\maketitle              
\begin{abstract}
Recent applications pose requirements of both cross-domain knowledge transfer and model compression to machine learning models due to insufficient training data and limited computational resources.
In this paper, we propose a new knowledge distillation model, named Spirit Distillation ($SD$), which is a model compression method with multi-domain knowledge transfer.
The compact student network mimics out a representation equivalent to the front part of the teacher network, through which the general knowledge can be transferred from the source domain (teacher) to the target domain (student).
To further improve the robustness of the student, we extend $SD$ to Enhanced Spirit Distillation ($ESD$) in exploiting a more comprehensive knowledge by introducing the proximity domain which is similar to the target domain for feature extraction.
Results demonstrate that our method can boost $mIOU$ and high-precision accuracy by 1.4\% and 8.2\% respectively with 78.2\% segmentation variance, and can gain a precise compact network with only 41.8\% FLOPs.

\end{abstract}
%
%
%

\section{Introduction}
Recent applications, such as self-driving cars and automated delivery robots, present the requirement of light-weight models due to limited computational resources as well as the real-time demand for recognition.
At the same time, as such applications often suffer from inadequate training data \cite{rare-situations,lackdata}, the introduction of cross-domain knowledge is urgently needed. Model compression \cite{purning,quan,tensordis}, which compress the formed network in
the back-end at the cost of a low loss of accuracy, and few-shot learning \cite{init1,init2,finetune}, which reduces the dependence of models on data through prior knowledge transfer, are presented to address these problems.

Among the various approaches, knowledge distillation and Fine-tuning-based Transfer Learning (FFT) are respectively considered as the most commonly used techniques for model compression and few-shot learning, and remarkable progress has been made in recent years \cite{hinton,fitnet,xie,liu1,finetune}. However, these methods can only solve one of these two problems, and so far there has been no study to combine them.

In our work, we pioneer cross-domain knowledge transfer under the framework of feature-based knowledge distillation \cite{kd-sum}, and introduce the Spirit Distillation ($SD$).
Different from previous approaches, $SD$ adopts the teacher and the student networks that address problems in different domains (source and target domain, respectively). The performance of the student network is improved by exploiting the potential to extract general features with cumbersome backbone discarded through general knowledge transfer from the source domain.
In addition, a more comprehensive general features extraction knowledge is transferred by extending $SD$ to Enhanced Spirit Distillation ($ESD$). 
By introducing extra data from the proximity domain which is similar to the target domain as general feature extraction materials, the student network can learn richer and more complete knowledge and achieve a more stable performance after fine-tuning.

In general, our contributions can be summarized as follows:
\begin{itemize}
	\item
We apply knowledge distillation to both model compression and few-shot learning and propose the Spirit Distillation ($SD$). Through general feature extraction knowledge transfer, the compact student network is able to learn an effective representation based on the front part of the teacher network.
	\item
We extend $SD$ to Enhanced Spirit Distillation ($ESD$). By introducing the proximity domain to achieve richer supervised intermediate representation, more complete knowledge can be learned by the student network, so that robustness of the student network can be significantly boosted.
	\item
Experiments on Cityscapes \cite{cityscapes} semantic segmentation with the prior knowledge transferred from COCO2017 \cite{coco} and KITTI \cite{kitti} demonstrate that: 
\begin{itemize}
	\item
	Spirit Distillation can significantly improve the performance of the student network (by 1.8\% mIOU enhancement) without enlarging the parameter size. 
	\item
	Enhanced Spirit Distillation can reinforce the robustness of the student network (8.2\% high-precision accuracy boosting and 21.8\% segmentation variance reduction) with comparable segmentation results attained.
\end{itemize}

\end{itemize}

\section{Related Work}
\subsection{Knowledge Distillation}
Knowledge distillation researches on the technical means of training compact student network with the prompt of cumbersome teacher network. Previous works can be mainly classified into logit-based distillation \cite{hinton,peng2019correlation,lkd} and feature-based distillation \cite{he,liu1,liu2,fitnet}, which transfer the knowledge from different stages of the teacher network to improve the performance of the student network. Pioneering works on knowledge distillation bases on logits tranfer \cite{hinton}, which adopt a weighted average of soft and hard labels as supervisory information for student network training. Subsequent works begin to focus on transferring intermediate representation of the teacher network, like FitNet stage-wise training \cite{fitnet}, knowledge adaptation \cite{he}, and structured knowledge distillation \cite{liu1}, hoping that the student network learns an effective representation based on the front part of the teacher with much fewer FLOPs.

\subsection{Few-shot Learning}
Few-shot learning provides a solution to the problems in scenarios with insufficient data, utilizing prior knowledge like understanding of the dataset or models trained on other datasets to reduce the dependence of machine learning models on data \cite{few-survey}. Existing few-shot learning methods based on data augmentation \cite{da1,da2,da3}, metric learning \cite{ml1,ml2,ml3}, and initialization \cite{init1,init2,init3} ameliorate the models in terms of supervised empirical growth, hypothesis space reduction, and initial parameter setting, respectively, so as to enhance the generalization ability of the models under the premise of inadequate training data.

\subsection{Fine-tuning-based Transfer Learning}
Fine-tuning-based transfer learning \cite{finetune} proposes to splice the problem-specific feature analysis part to the first few layers of a heavy network pretrained on a large-scale dataset, and train the constructed network under the condition of freezing the weights of the front part and do further fine-tuning. Since the layers transfered from the cubersome network are able to extract features with universal properties (i.e. general features), the trained-out network tends to be of great generalization capabilities.

\begin{figure}[!t]
	\centering
	\includegraphics[width=4.65in]{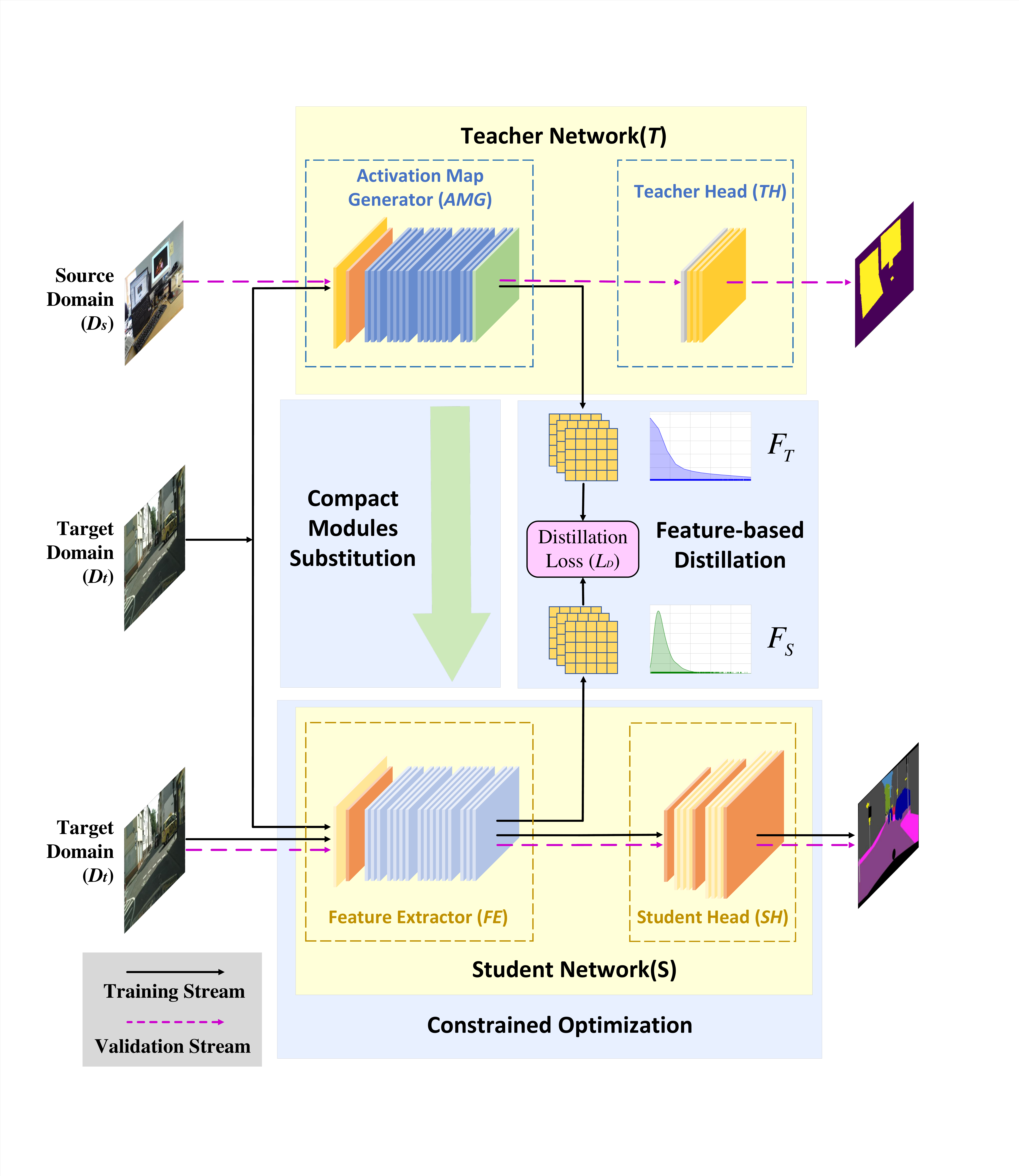}
		\caption{Framework of Spirit Distillation. The teacher network is pretrained on the source domain ($D_s$), while the student network is used to solve the problem on the target domain ($D_t$), which is with insufficient data. Three main steps are conducted: (1) Construct the student network by compact modules substitution and designing compact SH. (2) Learn a teacher-front-based representation utilizing feature-based knowledge distillation. (3) Perform constrained optimization on the student network.}
\end{figure}
\section{Approach}
\subsection{Framework of Spirit Distillation}
The basic framework of $SD$ is similar to feature-based knowledge distillation \cite{kd-sum}, which introduces both the teacher network ($T$) and the student network ($S$) in the training procedure, as shown in Fig. 1.
The teacher network adopts state-of-art architecture with pre-trained weights, and the student is compact and efficient.
This scheme of knowledge distillation allows the student to optimize by minimizing the distillation losses ($L_D$) between the hidden layer output features of the teacher ($F_T$) and the student ($F_S$), through which the student can learn a rich teacher-based intermediate representation. 
The optimization objective is defined as:
\begin{equation}
	W_S^{front} \gets \arg\min {L_D}({F_T},{F_S})
\end{equation}
thereout, we learn the weights of the front part of the student ($W_S^{front}$) from its teacher.

Unlike previous knowledge distillation methods, the teacher network and the student network in this paper are solving problems in different domains.
Our teacher is pretrained on the source domain ($D_s$), and the student is trained on the target domain ($D_t$).
Just as there is a huge gap in sample size and scenarios between $D_t$ and $D_s$, our goal is to improve the performance of $S$ on ${D_t}$ to the greatest extent, with powerful knowledge transferred from the representation of $T$ learned from ${D_s}$.
As shown in Fig. 1, $SD$ is conducted according to the following three steps:
\begin{itemize}
	\item
	Construct the student network by compact modules substitution and designing compact $SH$.
	\item
	Learn a teacher-front-based representation utilizing feature-based knowledge distillation.
	\item
	Perform constrained optimization on the student network.
\end{itemize}

Moreover, $ESD$ introduces the proximity domain ($D_p$) that is similar to ${D_t}$ and adopts data in $D_p$ as feature extraction materials, providing richer knowledge to enhance the distillation effect.

\subsection{Spirit Distillation}
\subsubsection{Student Network Construction}
Given a bulky pretrained teacher network $T$, we divide it into two parts according to the deviation between ${D_s}$ and ${D_t}$.
In this way, we gain the activation map generator ($AMG$) which is the first part, and the teacher head ($TH$) to be the second one.
Obtaining the $S$'s feature extractor ($FE$) by replacing the convolutional layers of $AMG$ with compact modules (e.g. group convolution \cite{alex}) to prepare the ground for efficient feature extraction. 
By designing the efficient feature analysis part for $S$ (i.e., student head, denoted as $SH$) and stacking the part after $FE$, the final $S$ is obtained.
As such, the inference cost of $S$ is much cheaper than that of $T$, and $FE$ has the potential to extract general features just like $AMG$ with even stronger generalization capability due to the smaller parameter size.
\subsubsection{Feature-based Distillation}
We input images of ${D_t}$ into $AMG$ and gain their general features (i.e., the output of the $AMG$, denoted as ${y^{AMG}}$). 
Suppose that the general features extracted by $AMG$ are “spirit” of the $T$'s representation for general feature extraction. 
These general features are less relevant to a specific domain and a particular network architecture compared with the hidden layer output of bulky networks converged on only the $D_t$'s training data. 
The rich semantic information of “spirit” for supervision is helpful knowledge to guide $FE$ to optimize toward extracting useful features for $D_t$.
As a result, we take ${y^{AMG}}$ as the optimization objective of the feature extractor ($FE$, whose output is denoted as ${y^{FE}}$) and transfer the “spirit” by minimizing the distillation loss ($L_D$).

\subsubsection{Constrained Optimization}
After transferring the knowledge from the teacher network, further optimization is required for precise prediction.
We first train $S$ (training loss function denoted as $L_P$) with a frozen $FE$, followed by a small learning rate optimization for the overall weights, to preserve the prior knowledge in the representation of $FE$ to the greatest extent.

\begin{figure}[!t]
	\centering
	\includegraphics[width=4.65in]{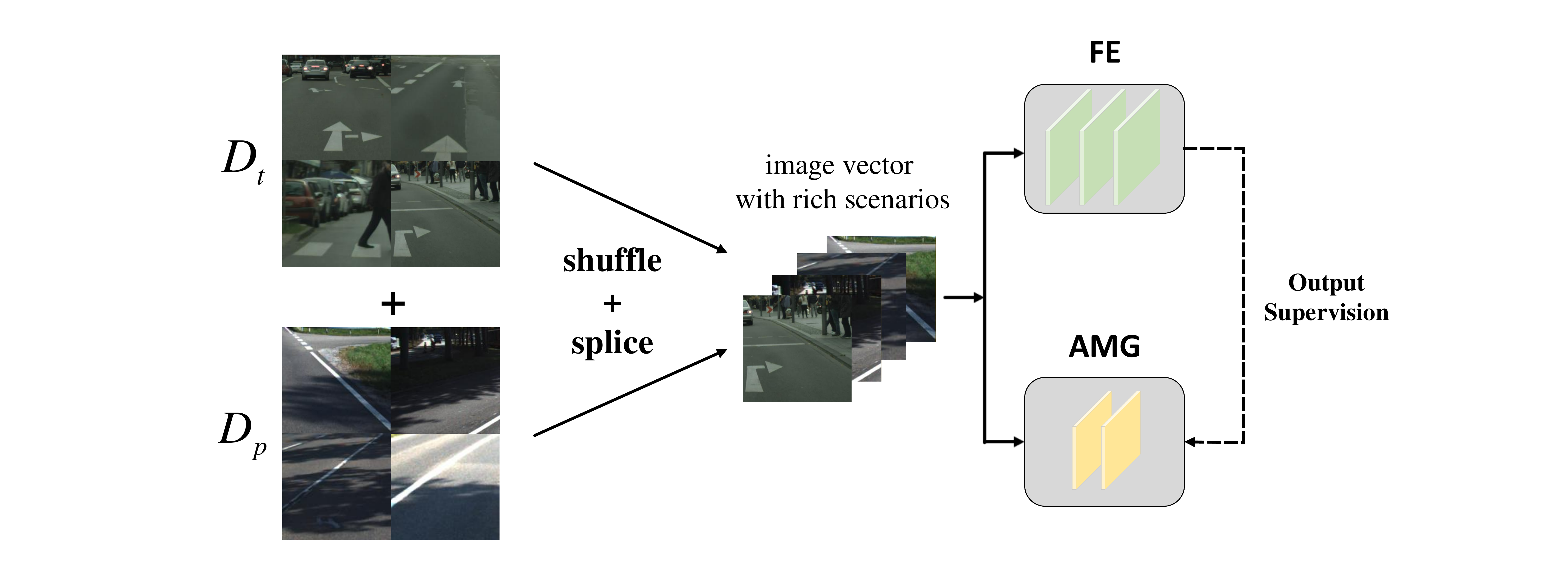}
	\caption{The alternative strategy for images input in $ESD$. Introduce the images of the proximity domain, and shuffle them with target domain images as feature extraction materials to provide richer knowledge.}
\end{figure}

\subsection{Enhanced Spirit Distillation}
Since the dataset of $D_t$ is largely undersampled from real scenes, the required diversity general features cannot be fully obtained by simply reinterpreting the images of ${D_t}$, which leads to the incomplete nature of the knowledge transferred.

Fortunately, feature representation knowledge learned from a particular dataset tends to work well for similar domains. 
Therefore, introducing a large-scale dataset of $D_p$ for feature extraction can prevent the feature extractor from overfitting to the little general features of $D_t$.
Moreover, $D_p$ may implicitly provide richer information of scenarios, and can compensate for the problem of insufficient data on ${D_t}$.

Based on the assumptions above, we extend $SD$ in terms of data inputting by shuffling ${D_t}$ and $D_p$ images together, extracting their features, and allowing the student network to imitate.
This method, shown in Fig. 2, expects to be executed in substitution with the input of $D_t$ images during the feature-based distillation process, and we name the newly integrated transferring and training scheme Enhanced Spirit Distillation ($ESD$).

\subsection{Formal Description of Enhanced Spirit Distillation}
Algorithm 1 provides a formal description of the overall procedure of Enhanced Spirit Distillation. The algorithm takes the weights of pre-trained teacher network ${W^T}$ (whose weights of $AMG$ part corresponds to ${W^{AMG}}$), the weights of randomly initialized student network ${W^S}$ (whose weights of $FE$ and $SH$ parts correspond to ${W^{FE}}$ and ${W^{ST}}$), the distillation loss $L_D$, the prediction loss $L_P$, the target domain dataset ${D_t}$, the proximity domain dataset ${D_p}$, and the optimizer $optimizer_i$ of the $i^{th}$ stage of training (with hyper settings) as inputs, and takes trained ${W^S}$ as output. Define ${W_1},{W_2},...{W_k}$ as the weights of layers $\{ {W_1},{W_2},...{W_k}\}$, ${\{ {W_1},{W_2},...{W_k}\} ^X}$ as the output of data $X$ (whose label is denoted as ${y^X}$) after the transformation operation of each layer, and $W^{k*}$ as the result of a certain iteration update of ${W^k}$.

\begin{algorithm}
	\caption{Enhanced Spirit Distillation}
	\KwIn{${W^T}$(${W^{AMG}}$), ${W^S}$(${W^{FE}}$, ${W^{ST}}$), $L_D$, $L_P$, ${D_t}$, ${D_p}$, $optimize{r_i}$}
	\KwOut{Trained ${W^S}$}
	\textbf{Stage 1: Feature-based Distillation} 

	\While{$FE$ not convergence}{
		${X_1}\gets shuffle\_select({D_p},{D_t})$\;
		${W^{FE*}}\gets {W^{FE}} - optimize{r_1}(\nabla {L_D}({\{ {W^{AMG}}\} ^{{X_1}}},{\{ {W^{FE}}\} ^{{X_1}}}))$\;
	}
	\textbf{Stage 2: Frozen Training}

	\While{$SH$ not convergence}{
		${X_2}, y^{X_2}\gets random\_choice({D_t})$\;
		${W^{SH*}} \gets {W^{SH}} - optimize{r_2}(\nabla {L_P}({\{ {W^S}\} ^{{X_2}}},y^{X_2}))$\;
	}
	\textbf{Stage 3: Fine-tuning}

	\While{$S$ not convergence}{
		${X_3}, y^{X_3}\gets random\_choice({D_t})$\;
		${W^{S*}}\gets {W^S} - optimize{r_3}(\nabla {L_{\rm{P}}}({\{ {W^S}\} ^{{X_3}}},y^{X_3}))$\;
	}
\end{algorithm}

\section{Experiments}
\subsection{Datasets}
We introduce COCO2017 \cite{coco}, Cityscapes \cite{cityscapes} and KITTI \cite{kitti} in our experiments, whose main properties, roles, and preprocessing methods are shown in Table 1. The subset of COCO2017 that contains the same class as Pascal VOC \cite{pascal} is used to pretrain the teacher network\footnote[1]{The pretrained weights of the teacher network are downloaded from \url{download.pytorch.org/models/deeplabv3\_resnet50\_coco-cd0a2569.pth}}. Only the first 64 images of Aachen in Cityscapes (denoted as Cityscapes-64) is chosen for feature-based distillation and constrained optimization. What's more, the images in KITTI are randomly shuffled with Cityscapes-64 ones in the feature-based distillation process when $ESD$ is adopted.


\begin{table}[!t]
	\centering
	\setlength{\abovecaptionskip}{15pt}%
	\setlength{\belowcaptionskip}{10pt}%
	\caption{The main properties, roles, and preprocessing methods of the datasets adopted in this paper. RC, RHF, MP, RZ in the table are abbreviations of random cropping, random horizontal flipping, maximum pooling and resizing, respectively.}
	\begin{tabular}{l|c|c|c|c|l}
		\hline
		\multicolumn{1}{c|}{Dataset} & \multicolumn{1}{l|}{Volume} & Resolution     & Scenario                            & \multicolumn{1}{c|}{Domain} & Preprocess                                                \\ \hline
		COCO2017 \cite{coco}                     & 100K+                       & -              & \multicolumn{1}{l|}{Common Objects} & Source                      & \multicolumn{1}{c}{-}                                     \\ \hline
		Cityscapes-64 \cite{cityscapes}                   & 64                        & 2048*1024      & \multirow{2}{*}{Road Scenes}        & Target                      & \begin{tabular}[c]{@{}l@{}}RC, RHF,\\ MP\end{tabular}     \\ \cline{1-3} \cline{5-6} 
		KITTI \cite{kitti}                        & 15K                         & About 1224*370 &                                     & Proximity                   & \begin{tabular}[c]{@{}l@{}}RZ, RC,\\ RHP, MP\end{tabular} \\ \hline
	\end{tabular}
\end{table}

\subsection{Network Architecture}
We adopt DeepLabV3 \cite{dv3} (resnet-50 \cite{cf1} backbone, pretrained on COCO2017) as the teacher network. To construct the feature extractor, we adopt the teacher's backbone with all of the convolutional layers replaced by group convolutions \cite{alex}, each group being the greatest common factor of the number of input and output channels. The student head is constructed by replacing the $ASPP$ and subsequent layers with a SegNet-like \cite{segnet} decoder structure, i.e., two groups of $3 \times (Conv+BN+ReLU)$ stack with bilinear up-sampling modules to achieve resolution increment and pixel-level classification. The convolution layers of the decoder also adopt group convolutions in the same setup as that adopted in the construction of $FE$.

\subsection{Implement Details}
\subsubsection{Basic Setup}
Experiments on binary segmentation on Cityscapes-64 are conducted to distinguish roads and backgrounds. Mean square error and pixel-average cross-entropy are taken as $L_D$ and $L_P$, respectively.

\subsubsection{Metrics}
We adopt mean intersection over union ($mIOU$) as the index to measure the segmentation effect, the size of parameters and floating point operations in measuring the compactness and inference efficiency of the network; the prediction variance and high-precision segmentation accuracy (considered to be segmented properly when $mIOU$>75\%, denoted as $(HP \rule[2.95pt]{0.1cm}{0.05em} Acc)$) in evaluating the robustness of the model.

\subsubsection{Hyper-parameter Settings}
We take a comparison experiment on whether or not to adopt $SD$ method.
We also employ the $FTT$ on the network that stacks $AMG$ and the $SH$ (denoted as constructed teacher network ($CT$)), and the former part is frozen in the first training stage.
We directly train the student network using stochastic gradient descent ($SGD$) with momentum $0.9$ and learning rate ${10^{ - 2}}$. For distillation process, a learning rate of $3 \times {10^{ - 3}}$ and a momentum of $0.99$ are adopted until convergence.
Constrained optimization requires freezing weights of $FE$. The training of the remaining portion adopts a momentum of $0.9$ with a learning rate of ${10^{ - 2}}$.
Further fine-tuning sets the learning rate of the entire network to $5 \times {10^{ - 5}}$ and the momentum to $0.99$. l2 weight penalty is adopted in all cases, with a decay constant of $3 \times {10^{ - 3}}$. 
Moreover, a data enhancement scheme with random cropping ($512 \times 512$) and random horizontal flipping is adopted, with max pooling (kernel\_size=2) adopted before input. 
To validate $ESD$, we set up a series of different scales to control the ratio $r$ that the number of input ${D_t}$ images to that of ${D_p}$ during distillation, and conduct the experiments separately. 
The ${D_t}$ images are preprocessed in the same way adopted for $SD$, except that they were previously resized to $2448 \times 740$ before cropping. All the preprocessing schemes for images in different datasets are shown in Table 1.

%
%



\begin{table}[!t]
	\centering
	\setlength{\abovecaptionskip}{15pt}%
	\setlength{\belowcaptionskip}{10pt}%
	\caption{The performance on the Cityscapes dataset in comparison with regular training and FTT. All the networks are trained only on the first 64 images of Cityscapes.}
	\begin{tabular}{lcccc}
		\hline
		\multicolumn{1}{c}{Method}                      & \multicolumn{1}{l}{GFLOPs} & \multicolumn{1}{l}{Param(M)} & \multicolumn{1}{l}{mIOU(\%)} & \multicolumn{1}{l}{HP-Acc(\%)} \\ \hline
		CT (FFT \cite{finetune}, without fine-tuning) & 405.7                      & 23.6                         & 62.6                         & 1.4                            \\ \hline
		CT (FFT \cite{finetune}, with fine-tuning)    & 405.7                      & 23.6                         & 58.9                         & 2.6                            \\ \hline
		S                           & \textbf{169.4}                      & \textbf{9.5}                          & 81.7                         & 81.2                           \\ \hline
		Ours: S (SD)                        & \textbf{169.4}                      & \textbf{9.5}                          & \textbf{83.5}                         & \textbf{84.6}                           \\ \hline
	\end{tabular}
\end{table}

\begin{table}[!t]
	\centering
	\setlength{\abovecaptionskip}{15pt}%
	\setlength{\belowcaptionskip}{10pt}%
	\caption{Comparison of the results on regular training, Spirit Distillation, and Enhanced Spirit Distillation with different $r$ values adopted.}
	\begin{tabular}{clccc}
		\hline
		\multicolumn{2}{c}{Method}                                 & mIOU(\%)      & HP-Acc(\%)    & Var($10^{-3}$)     \\ \hline
		\multicolumn{2}{c}{S}                                      & 81.7          & 81.2          & 5.77          \\ \hline
		\multicolumn{2}{c}{Ours: S (SD)}                             & \textbf{83.5} & 84.6          & 5.70          \\ \hline
		\multicolumn{1}{l}{} & r=10.0 & 82.2          & 85.4          & 5.12          \\ \cline{2-5} 
		\multicolumn{1}{l}{}                             & r=5.0  & 81.9          & 85.6          & 5.20          \\ \cline{2-5} 
		\multicolumn{1}{l}{}                             & r=3.0  & 82.2          & 84.6          & 5.21          \\ \cline{2-5} 
		\multicolumn{1}{l}{Ours: S (ESD)}                             & r=1.0  & 82.3          & 84.2          & \textbf{4.52} \\ \cline{2-5} 
		\multicolumn{1}{l}{}                             & r=0.5  & \textbf{83.3} & \textbf{89.2} & 4.71          \\ \cline{2-5} 
		\multicolumn{1}{l}{}                             & r=0.2  & 82.8          & \textbf{89.0} & \textbf{4.48} \\ \cline{2-5} 
		\multicolumn{1}{l}{}                             & r=0    & \textbf{83.1} & \textbf{89.4} & \textbf{4.51} \\ \hline
	\end{tabular}
\end{table}

\subsection{Results}
We display the segmentation results of the student network ($S$) trained with $SD$ along with the results of that trained follows regular training scheme and $FTT$ in Table 2. To validate the effectiveness of $ESD$, we respectively train $S$ under different $r$ settings and obtain the results shown in Table 3. We also plot the comparison of $mIOU$-$GFLOPs$, $Var$-$(HP \rule[2.95pt]{0.1cm}{0.05em} Acc)$, and $mIOU$-$r$ with different training settings, and calculated the distribution of segmentation effects when regular training, $SD$, and $ESD$ are adopted. (see Fig. 3)

The following conclusions can be drawn from Tables 2, 3 and Fig 3, 4.

\begin{figure}[!h]
	\centering
	\includegraphics[width=4.7in]{four.pdf}
	\caption{Comparison of $mIOU$-$GFLOPs$, $Var$-$(HP \rule[2.95pt]{0.1cm}{0.05em} Acc)$, and $mIOU$-$r$ with different training settings, along with the distribution of segmentation effects when regular training, $SD$, and $ESD$ are adopted.}
\end{figure}

\begin{figure}[!h]
	\centering
	\includegraphics[width=5.0in]{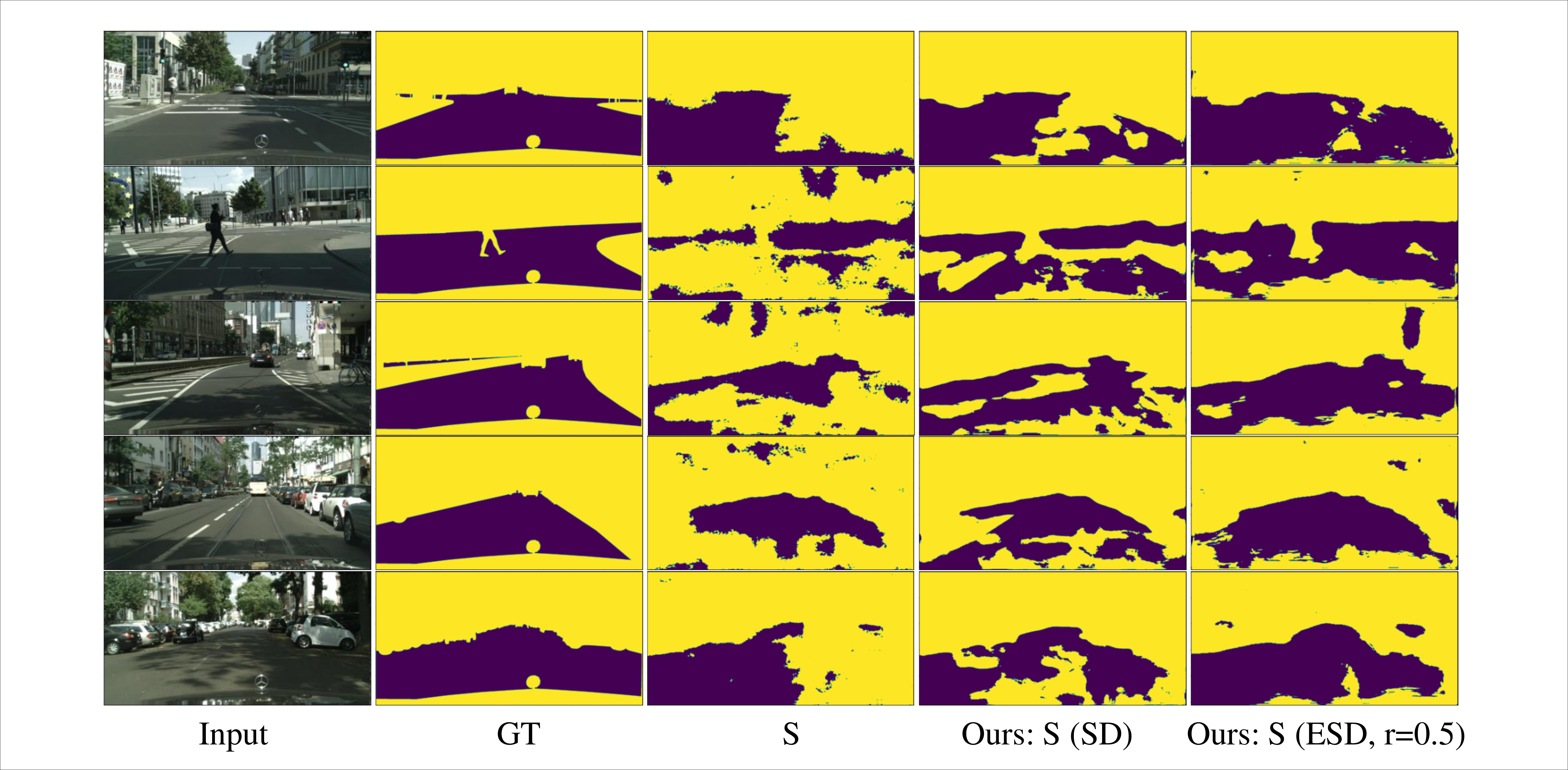}
	\caption{Comparison of segmentation results on Cityscapes-64. (1) Input images. (2) Ground truth. (3) The segmentation results of the student network. (4) The segmentation results of the student network that adopts Spirit Distillation. (5) The result of the student network that adopts Enhanced Spirit Distillation, with $r$ value set to be 0.5.}
\end{figure}

\begin{itemize}
	\item
	
	The $S$ using $SD$ outperforms normally trained $S$ as well as the fine-tune-transferred $CT$. In addition, with inference efficiency significantly improved, $SD$ also prevents the final network from over-fitting due to the large network size as well as under-fitting for the sake of freezing weights. (Table 2, Fig. 3(a))
	\item
	The segmentation effect of $S$ is improved with either $SD$ or $ESD$ adopted. The $S$ trained with $ESD$ can perform splendid predictions in more cases, and the proportion of very poor results is significantly reduced. Hence, $ESD$ can improve the robustness of $S$ to a great extent. (Fig. 3(b))
	\item
	The effectiveness of introducing $D_t$ is easily demonstrated as the final obtained $S$ tends to gain a higher $HP \rule[2.95pt]{0.1cm}{0.05em} Acc$ as well as $mIOU$ when the value of $r$ is set small, i.e., the proximity domain accounts for a larger proportion of the images used for feature extraction. (Table 2, Fig. 3(c))
	\item
	$ESD$ effectively improves the $HP \rule[2.95pt]{0.1cm}{0.05em} Acc$ while keeping the variances small numbers. The comprehensive learning of the general features extracted from $T$ helps prevent unstable prediction and enhances robustness. (Fig. 3(d))
	\item
	The validity of our methods in cross-domain knowledge transfer and robustness improvement under complex scenarios is easily confirmed, as the comparison shown in Fig. 4. The obvious finding is that the segmentation results of the undistilled $S$ are rather unsatisfactory for the shadow parts of the images. After adopting $SD$, the new $S$ has improved the segmentation results of these parts, which is able to distinguish the road scene from the shadows partially. Adopting the $ESD$ method on top of this, the $S$ would capture the global representation for shadow segmentation more completely and can distinguish the road part with shadows of the images more as a whole. (Fig. 4)
\end{itemize}

\section{Conclusion}
In order to introduce cross-domain knowledge while acquiring compressed models, a novel knowledge distillation method is proposed, which allows student networks to simulate part of the teacher's representation by transferring general knowledge from the large-scale source domain to the student network. 
To further boost the robustness of the student network, we introduce the proximity domain as the source of general feature extraction knowledge during feature-based distillation process.
Experiments demonstrate that our methods can effectively achieve cross-domain knowledge transfer and significantly boost the performance of compact models even with insufficient training data.

Future works will include extending our approach to other visual applications and conducting domain transformation to feature extraction materials using approaches like conditional generative adversarial network training.

%
%
%
 \bibliographystyle{splncs04}
 \bibliography{references}

\begin{thebibliography}{10}
\providecommand{\url}[1]{\texttt{#1}}
\providecommand{\urlprefix}{URL }
\providecommand{\doi}[1]{https://doi.org/#1}

\bibitem{segnet}
Badrinarayanan, V., Kendall, A., Cipolla, R.: Segnet: A deep convolutional
  encoder-decoder architecture for image segmentation. IEEE transactions on
  pattern analysis and machine intelligence  \textbf{39}(12),  2481--2495
  (2017)

\bibitem{ml3}
Bertinetto, L., Henriques, J.F., Torr, P.H., Vedaldi, A.: Meta-learning with
  differentiable closed-form solvers. arXiv preprint arXiv:1805.08136  (2018)

\bibitem{lackdata}
Biasetton, M., Michieli, U., Agresti, G., Zanuttigh, P.: Unsupervised domain
  adaptation for semantic segmentation of urban scenes. In: Proceedings of the
  IEEE/CVF Conference on Computer Vision and Pattern Recognition Workshops.
  pp.~0--0 (2019)

\bibitem{dv3}
Chen, L.C., Papandreou, G., Schroff, F., Adam, H.: Rethinking atrous
  convolution for semantic image segmentation. arXiv preprint arXiv:1706.05587
  (2017)

\bibitem{da3}
Chen, Z., Fu, Y., Chen, K., Jiang, Y.G.: Image block augmentation for one-shot
  learning. In: Proceedings of the AAAI Conference on Artificial Intelligence.
  vol.~33, pp. 3379--3386 (2019)

\bibitem{da2}
Chen, Z., Fu, Y., Zhang, Y., Jiang, Y.G., Xue, X., Sigal, L.: Multi-level
  semantic feature augmentation for one-shot learning. IEEE Transactions on
  Image Processing  \textbf{28}(9),  4594--4605 (2019)

\bibitem{cityscapes}
Cordts, M., Omran, M., Ramos, S., Rehfeld, T., Enzweiler, M., Benenson, R.,
  Franke, U., Roth, S., Schiele, B.: The cityscapes dataset for semantic urban
  scene understanding. In: Proceedings of the IEEE conference on computer
  vision and pattern recognition. pp. 3213--3223 (2016)

\bibitem{tensordis}
Denton, E., Zaremba, W., Bruna, J., LeCun, Y., Fergus, R.: Exploiting linear
  structure within convolutional networks for efficient evaluation. arXiv
  preprint arXiv:1404.0736  (2014)

\bibitem{pascal}
Everingham, M., Van~Gool, L., Williams, C.K., Winn, J., Zisserman, A.: The
  pascal visual object classes (voc) challenge. International journal of
  computer vision  \textbf{88}(2),  303--338 (2010)

\bibitem{init1}
Finn, C., Abbeel, P., Levine, S.: Model-agnostic meta-learning for fast
  adaptation of deep networks. In: International Conference on Machine
  Learning. pp. 1126--1135. PMLR (2017)

\bibitem{kitti}
Geiger, A., Lenz, P., Urtasun, R.: Are we ready for autonomous driving? the
  kitti vision benchmark suite. In: 2012 IEEE Conference on Computer Vision and
  Pattern Recognition. pp. 3354--3361. IEEE (2012)

\bibitem{da1}
Hariharan, B., Girshick, R.: Low-shot visual recognition by shrinking and
  hallucinating features. In: Proceedings of the IEEE International Conference
  on Computer Vision. pp. 3018--3027 (2017)

\bibitem{cf1}
He, K., Zhang, X., Ren, S., Sun, J.: Deep residual learning for image
  recognition. In: Proceedings of the IEEE conference on computer vision and
  pattern recognition. pp. 770--778 (2016)

\bibitem{he}
He, T., Shen, C., Tian, Z., Gong, D., Sun, C., Yan, Y.: Knowledge adaptation
  for efficient semantic segmentation. In: Proceedings of the IEEE/CVF
  Conference on Computer Vision and Pattern Recognition. pp. 578--587 (2019)

\bibitem{hinton}
Hinton, G., Vinyals, O., Dean, J.: Distilling the knowledge in a neural
  network. arXiv preprint arXiv:1503.02531  (2015)

\bibitem{alex}
Krizhevsky, A., Sutskever, I., Hinton, G.E.: Imagenet classification with deep
  convolutional neural networks. Advances in neural information processing
  systems  \textbf{25},  1097--1105 (2012)

\bibitem{coco}
Lin, T.Y., Maire, M., Belongie, S., Hays, J., Perona, P., Ramanan, D.,
  Doll{\'a}r, P., Zitnick, C.L.: Microsoft coco: Common objects in context. In:
  European conference on computer vision. pp. 740--755. Springer (2014)

\bibitem{liu1}
Liu, Y., Chen, K., Liu, C., Qin, Z., Luo, Z., Wang, J.: Structured knowledge
  distillation for semantic segmentation. In: Proceedings of the IEEE/CVF
  Conference on Computer Vision and Pattern Recognition. pp. 2604--2613 (2019)

\bibitem{liu2}
Liu, Y., Shu, C., Wang, J., Shen, C.: Structured knowledge distillation for
  dense prediction. IEEE Transactions on Pattern Analysis and Machine
  Intelligence  (2020)

\bibitem{purning}
Luo, J.H., Wu, J., Lin, W.: Thinet: A filter level pruning method for deep
  neural network compression. In: Proceedings of the IEEE international
  conference on computer vision. pp. 5058--5066 (2017)

\bibitem{peng2019correlation}
Peng, B., Jin, X., Liu, J., Li, D., Wu, Y., Liu, Y., Zhou, S., Zhang, Z.:
  Correlation congruence for knowledge distillation. In: Proceedings of the
  IEEE/CVF International Conference on Computer Vision. pp. 5007--5016 (2019)

\bibitem{rare-situations}
Pouyanfar, S., Saleem, M., George, N., Chen, S.C.: Roads: Randomization for
  obstacle avoidance and driving in simulation. In: Proceedings of the IEEE/CVF
  Conference on Computer Vision and Pattern Recognition Workshops. pp.~0--0
  (2019)

\bibitem{init3}
Ravi, S., Larochelle, H.: Optimization as a model for few-shot learning  (2016)

\bibitem{fitnet}
Romero, A., Ballas, N., Kahou, S.E., Chassang, A., Gatta, C., Bengio, Y.:
  Fitnets: Hints for thin deep nets. arXiv preprint arXiv:1412.6550  (2014)

\bibitem{init2}
Rusu, A.A., Rao, D., Sygnowski, J., Vinyals, O., Pascanu, R., Osindero, S.,
  Hadsell, R.: Meta-learning with latent embedding optimization. arXiv preprint
  arXiv:1807.05960  (2018)

\bibitem{ml1}
Snell, J., Swersky, K., Zemel, R.S.: Prototypical networks for few-shot
  learning. arXiv preprint arXiv:1703.05175  (2017)

\bibitem{ml2}
Sung, F., Yang, Y., Zhang, L., Xiang, T., Torr, P.H., Hospedales, T.M.:
  Learning to compare: Relation network for few-shot learning. In: Proceedings
  of the IEEE conference on computer vision and pattern recognition. pp.
  1199--1208 (2018)

\bibitem{lkd}
Tian, Y., Krishnan, D., Isola, P.: Contrastive representation distillation.
  arXiv preprint arXiv:1910.10699  (2019)

\bibitem{kd-sum}
Wang, L., Yoon, K.J.: Knowledge distillation and student-teacher learning for
  visual intelligence: A review and new outlooks. IEEE Transactions on Pattern
  Analysis and Machine Intelligence  (2021)

\bibitem{few-survey}
Wang, Y., Yao, Q., Kwok, J.T., Ni, L.M.: Generalizing from a few examples: A
  survey on few-shot learning. ACM Computing Surveys (CSUR)  \textbf{53}(3),
  1--34 (2020)

\bibitem{quan}
Wu, J., Leng, C., Wang, Y., Hu, Q., Cheng, J.: Quantized convolutional neural
  networks for mobile devices. In: Proceedings of the IEEE Conference on
  Computer Vision and Pattern Recognition. pp. 4820--4828 (2016)

\bibitem{xie}
Xie, J., Shuai, B., Hu, J.F., Lin, J., Zheng, W.S.: Improving fast segmentation
  with teacher-student learning. arXiv preprint arXiv:1810.08476  (2018)

\bibitem{finetune}
Yosinski, J., Clune, J., Bengio, Y., Lipson, H.: How transferable are features
  in deep neural networks? arXiv preprint arXiv:1411.1792  (2014)

\end{thebibliography}
\end{document}